\documentclass[11pt,a4paper]{article}
\usepackage[hyperref]{naaclhlt2018}
\usepackage{times}
\usepackage{latexsym}
\usepackage{paralist}
\usepackage{booktabs} 
\usepackage{amsfonts}
\usepackage{booktabs}
\usepackage{amsmath}
\usepackage{tabularx}
\usepackage{tabulary}
\usepackage{todonotes}
\usepackage{multirow}
\usepackage{url}
\usepackage{hyperref}

\aclfinalcopy 

\title{Zero-Shot Question Generation from Knowledge Graphs for Unseen Predicates and Entity Types}

\author{
  Hady Elsahar, Christophe Gravier, Frederique Laforest\\
  Universit\'e de Lyon\\
  Laboratoire Hubert Curien\\
  Saint-\'Etienne, France \\
  \texttt{\{hady.elsahar, christophe.gravier, frederique.laforest\}@univ-st-etienne.fr} \\
}

\begin{document}
\maketitle

\begin{abstract}

%
%
We present a neural model for question generation from knowledge base triples in a ``Zero-Shot" setup, that is generating questions for triples containing predicates, subject types or object types that were not seen at training time. 
Our model leverages triples occurrences in the natural language corpus in an encoder-decoder architecture, paired with an original part-of-speech copy action mechanism to generate questions. %
Benchmark and human evaluation show that our model sets a new state-of-the-art for zero-shot QG.
\end{abstract}

\section{Introduction}
%
%
%
Questions Generation (QG) from Knowledge Graphs is the task consisting in generating natural language questions given an input knowledge base (KB) triple~\cite{Serban16}.
QG from knowledge graphs has shown to improve the performance of existing factoid question answering (QA) systems either by dual training or by augmenting existing training datasets~\cite{Dong17,Khapra17}.
%
%
%
%
Those methods rely on large-scale annotated datasets such as SimpleQuestions~\cite{Bordes15}. 
In practice many of the predicates and entity types in KB are not covered by those annotated datasets. For example $75\%$ of Freebase predicates are not covered by SimpleQuestions.
%
So, one challenge for QG from knowledge graphs is to adapt to predicates and entity types that were \textit{not} seen at training time (aka Zero-Shot QG).
%
%
Ultimately, generating questions to predicates and entity types unseen at training time will allow QA systems to cover predicates and entity types that would not have been used for QA otherwise.

Intuitively, a human who is given the task to write a question on a fact offered by a KB, would read natural language sentences where the entity or the predicate of the fact occur, and build up questions that are aligned with what he reads from both a lexical and grammatical standpoint.
%
In this paper, we propose a model for Zero-Shot Question Generation that follows this intuitive process. 
In addition to the input KB fact, we feed our model with a set of textual contexts paired with the input KB triple through distant supervision.
Our model derives an encoder-decoder architecture, in which the encoder encodes the input KB triple, along with a set of textual contexts into hidden representations. 
Those hidden representations are fed to a decoder equipped with an attention mechanism to generate an output question. \\
%
%
In the Zero-Shot setup, the emergence of new predicates and new class types during test time requires new lexicalizations to express these predicates and classes in the output question. These lexicalizations might not be encountered by the model during training time and hence do not exist in the model vocabulary, or have been seen only few times not enough to learn a good representation for them by the model.
Recent works on Text Generation tackle the rare words/unknown words problem using copy actions~\cite{Luong15,Gulcehre16}: words with a specific position are copied from the source text to the output text -- although this process is blind to the role and nature of the word in the source text.
Inspired by research in open information extraction~\cite{Fader11} and structure-content neural language models~\cite{Kiros14}, in which part-of-speech tags represent a distinctive feature when representing relations in text, we extend these positional copy actions.
Instead of copying a word in a specific position in the source text, our model copies a word with a specific part-of-speech tag from the input text -- we refer to those as part-of-speech copy actions.
%
%
%
Experiments show that our model using contexts through distant supervision significantly outperforms the strongest baseline among six ($+2.04$ BLEU-4 score). %
Adding our copy action mechanism further increases this improvement ($+2.39$).
Additionally, a human evaluation complements the comprehension of our model for edge cases
; it supports the claim that the improvement brought by our copy action mechanism is even more significant than what the BLEU score suggests. 
%
%
%
%
\section{Related Work}
QG became an essential component in many applications such as education~\cite{HeilmanS10}, tutoring~\cite{graesser2004,evens2006} and dialogue systems~\cite{Shang15}. 
In our paper we focus on the problem of QG from structured KB and how we can generalize it to unseen predicates and entity types. 
%
\cite{SeylerYB15} generate quiz questions from KB triples. Verbalization of entities and predicates relies on their existing labels in the KB and a dictionary. 
\cite{Serban16} use an encoder-decoder architecture with attention mechanism trained on the SimpleQuestions dataset~\cite{Bordes15}.
\cite{Dong17} generate paraphrases of given questions to increases the performance of QA systems; paraphrases are generated relying on paraphrase datasets, neural machine translation and rule mining.
\cite{Khapra17} generate a set of QA pairs given a KB entity. 
They model the problem of QG as a sequence to sequence problem by converting all the KB entities to a set of keywords. 
None of the previous work in QG from KB address the question of generalizing to unseen predicates and entity types. \\
%
Textual information has been used before in the Zero-Shot learning. \cite{Socher13} use information in pretrained word vectors for Zero-Shot visual object recognition.
\cite{Levy17} incorporates a natural language question to the relation query to tackle Zero-Shot relation extraction problem.
%

%
Previous work in machine translation dealt with rare or unseen word problem problem for translating names and numbers in text. 
\cite{Luong15} propose a model that generates positional placeholders pointing to some words in source sentence and copy it to target sentence (\textit{copy actions}).
\cite{Gulcehre16,Gu16} introduce separate trainable modules for copy actions to adapt to highly variable input sequences, for text summarization. 
For text generation from tables, \cite{lebret2016} extend positional copy actions to copy values from fields in the given table.
For QG,~\cite{Serban16} use a placeholder for the subject entity in the question to generalize to unseen entities. 
Their work is limited to unseen entities and does not study how they can generalize to unseen predicates and entity types.
%
%

\begin{figure*}[h]
  \centering
  \small
  \def\svgwidth{.75\linewidth}
  \includegraphics[width=0.75\textwidth]{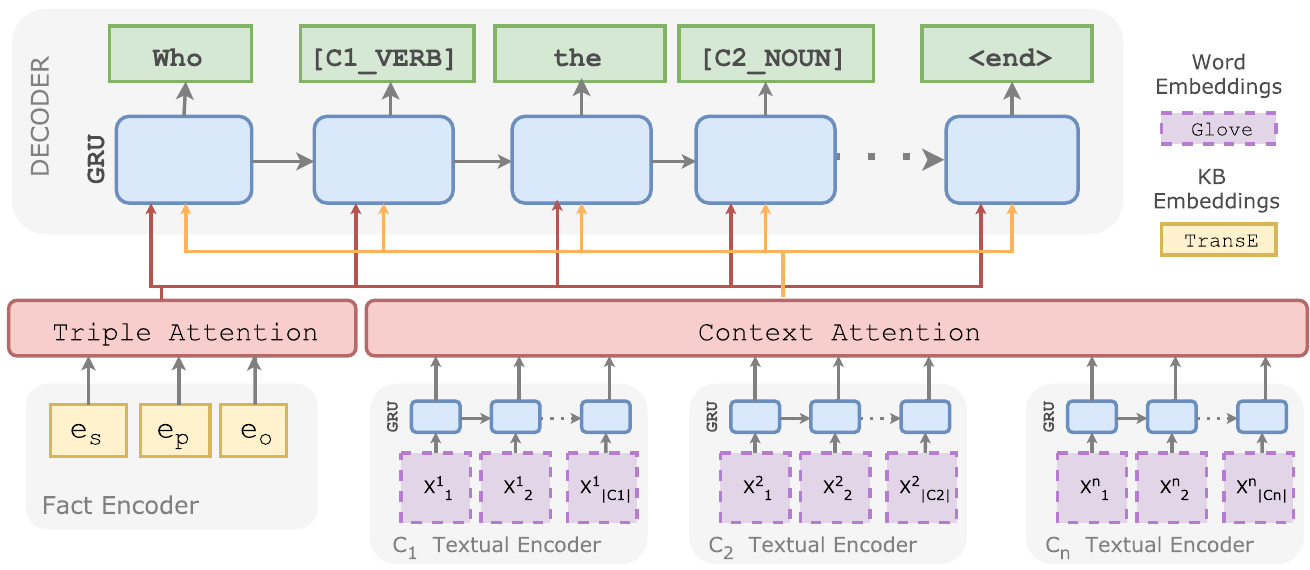}
  \caption{The proposed model for Question Generation. The model consists of a single fact encoder and $n$ textual context encoders, each consists of a separate GRU. At each time step $t$, two attention vectors generated from the two attention modules are fed to the decoder to generate the next word in the output question.}
  \label{fig:model}
\end{figure*}
%
\section{Model}
Let $F=\{s,p,o\}$ be the input fact provided to our model consisting of a subject $s$, a predicate $p$ and an object $o$, and $C$ be the set of textual contexts associated to this fact.
Our goal is to learn a model that generates a sequence of $T$ tokens 
$Y = y_1,y_2,\ldots,y_T$ representing a question about the subject $s$, where the object $o$ is the correct answer.
Our model approximates the conditional probability of the output question given an input fact $p(Y|F)$, to be the probability of the output question, given an input fact and the additional textual context $C$, modelled as follows:
\begin{align}
p(Y|F) &= \prod_{t=1}^{T}p(y_t|y_{<t}, F,C)
\end{align}
where $y_{<t}$ represents all previously generated tokens until time step $t$. %
Additional textual contexts are natural language representation of the triples that can be drawn from a corpus -- our model is generic to any textual contexts that can be additionally provided, though we describe in Section~\ref{subsec:textualcontexts} how to create such texts from Wikipedia.

Our model derives the encoder-decoder architecture of~\cite{Sutskever14,Bahdanau14} with two encoding modules: a feed forward architecture encodes the input triple (sec. \ref{sec:fact-encoder}) and a set of recurrent neural network (RNN) to encode each textual context (sec. \ref{sec:textual-encoder}). 
Our model has two attention modules~\cite{Bahdanau14}: one acts over the input triple and another acts over the input textual contexts (sec. \ref{sec:attention}).
The decoder (sec. \ref{sec:decoder}) is another RNN that generates the output question. At each time step, the decoder chooses to output either a word from the vocabulary or a special token indicating a copy action (sec. \ref{sec:copy-actions}) from any of the textual contexts.
%
\subsection{Fact Encoder}~\label{sec:fact-encoder}
Given an input fact $F=\{s,p,o\}$, let each of $e_s$, $e_p$ and $e_o$ be a 1-hot vectors of size $K$. 
The fact encoder encodes each 1-hot vector into a fixed size vector $h_s = \mathbf{E_f}\,e_s$, \enspace $h_p = \mathbf{E_f}\,e_p$ \enspace and $h_o = \mathbf{E_f}\,e_o$,
where $\mathbf{E_f} \in \mathbb{R}^{H_k \times K}$ is the KB embedding matrix, $H_k$ is the size of the KB embedding and $K$ is the size of the KB vocabulary. 
The \emph{encoded fact} $ h_f \in \mathbb{R}^{3H_k}$ represents the concatenation of those three vectors and we use it to initialize the decoder. 
\begin{align}\label{eq:fact-emb}
h_f = [h_s;\enspace h_p;\enspace h_o]    
\end{align}
%
Following~\cite{Serban16}, we learn $\mathbf{E_{f}}$ using \textit{TransE}~\cite{Bordes15}. We fix its weights and do not allow their update during training time. 
%
\subsection{Textual Context Encoder}~\label{sec:textual-encoder}
Given a set of $n$ textual contexts $ C = \{c_1, c_2, \ldots, c_n\ : c_j = (x_1^j, x_2^j,  \ldots, x^j_{|c_j|})\}$, where $x_i^j$ represents the 1-hot vector of the $i^{th}$ token in the $j^{th}$ textual context $c_j$, and $|c_j|$ is the length of the $j^{th}$ context.
We use a set of $n$ Gated Recurrent Neural Networks (GRU)~\cite{Cho2014} to encode each of the textual concepts separately:
\begin{align}\label{eq:gru-enc}
 h_i^{c_j} = GRU_j\left (\mathbf{E_c}\,x_{i}^{j},\enspace h_{i-1}^{c_j} \right )
\end{align}
where $h_i^{c_j} \in \mathbb{R}^{H_c}$ is the hidden state of the GRU that is equivalent to $x_i^j$ and of size $H_c$ . $\mathbf{E_{c}}$ is the input word embedding matrix.
The \emph{encoded context} represents the encoding of all the textual contexts; it is calculated as the concatenation of all the final states of all the encoded contexts:
\begin{align}\label{eq:context-emb}
    h_c &= [h^{c_1}_{|c_1|}; h^{c_2}_{|c_2|};\ldots; h^{c_n}_{|c_n|}].    
\end{align}
%
%
\subsection{Decoder} \label{sec:decoder}
For the decoder we use another GRU with an attention mechanism~\cite{Bahdanau14}, in which the decoder hidden state $s_t\in\mathbb{R}^{H_{d}}$ at each time step $t$ is calculated as:
\begin{align}
s_{t} &= z_t \circ s_{t-1} + (1-z_t) \circ \tilde{s}_{t} \enspace,
\end{align}
Where:
\small
\begin{align}
\tilde{s}_{t}  &= tanh\left(W E_{w}y_{t-1} + U [r_t \circ s_{t-1}] + A \, [a^f_t; a^c_t] \right) \\
z_t  &= \sigma\left(W_z\,E_{w}\,y_{t-1} + U_z \, s_{t-1} + A_z \, [a^f_t; a^c_t]  \right) \\
r_t  &= \sigma\left(W_r\,E_{w}\,y_{t-1} + U_r \, s_{t-1} + A_r \, [a^f_t; a^c_t]  \right)
\end{align}
\normalsize
%
$W,W_z,W_r \in \mathbb{R}^{m \times H_{d}}$, $U,U_z,U_r,A,A_z,A_r \in \mathbb{R}^{H_{d} \times H_{d}}$ are learnable parameters of the GRU. 
$E_w \in \mathbf{R}^{m \times V}$ is the word embedding matrix, $m$ is the word embedding size and $H_d$ is the size of the decoder hidden state. 
$a^f_t$, $a^c_t$ are the outputs of the fact attention and the context attention modules respectively, detailed in the following subsection.  \\
In order to enforce the model to pair output words with words from the textual inputs, we couple the word embedding matrices of both the decoder $E_w$ and the textual context encoder $E_c$ (eq.(\ref{eq:gru-enc})). 
We initialize them with GloVe embeddings~\cite{PenningtonSM14} and allow the network to tune them.\\
The first hidden state of the decoder $s_0 = [h_f;\,h_c]$ is initialized using a concatenation of the encoded fact~(eq.(\ref{eq:fact-emb})) and the encoded context~(eq.(\ref{eq:context-emb})) .\\
At each time step $t$, after calculating the hidden state of the decoder, the conditional probability distribution over each token $y_t$ of the generated question is computed as the $softmax(W_{o}\,s_t)$ over all the entries in the output vocabulary, $W_o \in \mathbb{R}^{H_d \times V}$ is the weight matrix of the output layer of the decoder.\\
%
\subsection{Attention} \label{sec:attention}
Our model has two attention modules: \\
\textbf{Triple attention} over the input triple to determine at each time step $t$ an attention-based encoding of the input fact $a^f_t \in \mathbb{R}^{H_k}$: 
\begin{align}
a^f_t = \alpha_{s,t} \, h_{s} + \alpha_{p,t} \, h_p + \alpha_{s,t} \, h_o   \enspace,
\end{align}
$\alpha_{s,t}, \alpha_{p,t}, \alpha_{o,t}$ are scalar values calculated by the attention mechanism to determine at each time step which of the encoded subject, predicate, or object the decoder should attend to. \\
\textbf{Textual contexts attention} over all the hidden states of all the textual contexts $a^c_t \in \mathbb{R}^{H_c}$:
\begin{align}
    a^c_t = \sum_{i=1}^{|C|} \sum_{j=1}^{|c_i|} \alpha_{t,j}^{c_i} \, h_j^{c_i}    \enspace,
\end{align}
$\alpha_{t,j}^{c_i}$ is a scalar value determining the weight of the $j^{th}$ word in the $i^{th}$ context $c^i$ at time step $t$. 
%

Given a set of encoded input vectors $I = \{h_1,h_2,...h_k\}$ and the decoder previous hidden state $s_{t-1}$, the attention mechanism calculates $\alpha_{t} = \alpha_{i,t},\ldots,\alpha_{k,t}$ as a vector of scalar weights, each $\alpha_{i,t}$ determines the weight of its corresponding encoded input vector $h_i$. 
\begin{align}
e_{i,t} &= \mathbf{v_a}^\top \enspace tanh(\mathbf{W_a \, s_{t-1} + U_a \, h_i}) \\
\alpha_{i,t} &= \frac{exp\left(e_{i,t}\right)}{\sum_{j=1}^{k} exp\left(e_{j,t}\right)} \enspace,
\end{align} where 
$\mathbf{v_a}, \mathbf{W_a}, \mathbf{U_a}$ are trainable weight matrices of the attention modules. 
It is important to notice here that we encode each textual context separately using a different GRU, but we calculate an overall attention over all tokens in all textual contexts: at each time step the decoder should ideally attend to only one word from all the input contexts.
%

\begin{table}[t]
\centering
\footnotesize 
\begin{tabular}{@{}ll@{}}
\midrule[1.1pt]
\multicolumn{2}{l}{What caused the \texttt{{[}C1\_NOUN{]}} of the \texttt{{[}C3\_NOUN{]}} \texttt{{[}S{]}} ?} \\ \midrule[1pt]
{\multirow{2}{*}{C1}} & \texttt{{[}S{]} \textbf{death} by {[}O{]}} \\                                           \cmidrule(l){2-2}
                      & \texttt{{[}S{]} \textbf{{[}C1\_NOUN{]}} {[}C1\_ADP{]} {[}O{]}} \\                    \midrule[1pt]
{\multirow{2}{*}{C2}} & {Disease} \\ \cmidrule(l){2-2} 
                      & \texttt{{[}C2\_NOUN{]}}               \\\midrule[1pt]
{\multirow{2}{*}{C3}} & {Musical \textbf{artist}}           \\ \cmidrule(l){2-2}
                      & \texttt{{[}C3\_ADJ{]}} \textbf{{\texttt{[}C3\_NOUN{]}}} \\ \bottomrule
\end{tabular}
\caption{An annotated example of part-of-speech copy actions from several input textual contexts (C1, C2, C3), the words or placeholders in bold are copied in the generated question}
\label{table:copyactions}
\end{table}
\subsection{Part-Of-Speech Copy Actions}\label{sec:copy-actions}
%
%
We use the method of \cite{Luong15} by modeling all the copy actions on the data level through an annotation scheme. 
This method treats the model as a black box, which makes it adaptable to any text generation model. 
Instead of using positional copy actions, we use the part-of-speech information to decide the alignment process between the input and output texts to the model. 
Each word in every input textual context is replaced by a special token containing a combination of its context id (e.g. \texttt{C1}) and its POS tag (e.g. \texttt{NOUN}).
Then, if a word in the output question matches a word in a textual context, it is replaced with its corresponding tag as shown in Table~\ref{table:copyactions}. \\
Unlike \cite{Serban16,lebret2016} we model the copy actions in the input and the output levels. 
Our model does not have the drawback of losing the semantic information when replacing words with generic placeholders, since we provide the model with the input triple through the fact encoder.
During inference the model chooses to either output words from the vocabulary or special tokens to copy from the textual contexts. 
In a post-processing step those special tokens are replaced with their original words from the textual contexts.
%

\section{Textual contexts dataset}
As a source of question paired with KB triples we use the SimpleQuestions dataset~\cite{Bordes15}. 
It consists of 100K questions with their corresponding triples from Freebase, and was created manually through crowdsourcing.
When asked to form a question from an input triple, human annotators usually tend to mainly focus on expressing the predicate of the input triple.
For example, given a triple with the predicate \texttt{fb:spacecraft/manufacturer} the user may ask \textit{"What is the manufacturer of \texttt{[S]} ?"}.
Annotators may specify the entity type of the subject or the object of the triple: \textit{"What is the manufacturer of the \textbf{spacecraft} \texttt{[S]}?"} or \textit{"Which \textbf{company} manufactures \texttt{[S]}?"}.
Motivated by this example we chose to associate each input triple with three textual contexts of three different types. 
The first is a phrase containing lexicalization of the predicate of the triple. 
The second and the third are two phrases containing the entity type of the subject and the object of the triple. 
In what follows we show the process of collection and preprocessing of those textual contexts.
\subsection{Collection of Textual Contexts}\label{subsec:textualcontexts}
We extend the set of triples given in the SimpleQuestions dataset by using the FB5M~\cite{Bordes15} subset of Freebase. As a source of text documents, we rely on Wikipedia articles.
\paragraph{Predicate textual contexts:}
In order to collect textual contexts associated with the SimpleQuestions triples, we follow the distant supervision setup for relation extraction~\cite{mintz_2009}.
The distant supervision assumption has been effective in creating training data for relation extraction and shown to be 87\% correct~\cite{riedel_2010} on Wikipedia text. \\
First, we align each triple in the FB5M KB to sentences in Wikipedia if the subject and the object of this triple co-occur in the same sentence. 
We use a simple string matching heuristic to find entity mentions in text\footnote{
We map Freebase entities to Wikidata through the Wikidata property P646, then we extract their labels and aliases. 
We use the Wikidata truthy dump:~\url{https://dumps.wikimedia.org/wikidatawiki/entities/}}. 
Afterwards we reduce the sentence to the set of words that appear on the dependency path between the subject and the object mentions in the sentence.
We replace the positions of the subject and the object mentions with \texttt{[S]} and \texttt{[O]} to the keep track of the information about the direction of the relation.
The top occurring pattern for each predicate is associated to this predicate as its textual context.
Table~\ref{table:example-dep} shows examples of predicates and their corresponding textual context.
\paragraph{Sub-Type and Obj-Type textual contexts:}
We use the labels of the entity types as the sub-type and obj-type textual contexts. 
We collect the list of entity types of each entity in the FB5M through the predicate \texttt{fb:type/instance}.
If an entity has multiple entity types we pick the entity type that is mentioned the most in the first sentence of each Wikipedia article. 
Thus the textual contexts will opt for entity types that is more natural to appear in free text and therefore questions. 
%
\begin{table}[t]
\centering
\footnotesize
\begin{tabular}{@{}l@{\hskip -0.1pt}l@{}}
\toprule
\textbf{Freebase Relation}                                                   & \textbf{Predicate Textual Context}                                                  \\ \midrule
person/place\_of\_birth                     & {[}O{]} is birthplace of {[}S{]}                 \\
currency/former\_countries         & {[}S{]} was currency of {[}O{]}                  \\
airline\_accident/operator               & {[}S{]} was accident for {[}O{]}                 \\
genre/artists                                & {[}S{]} became a genre of {[}O{]}                \\
risk\_factor/diseases                     & {[}S{]} increases likelihood of {[}O{]} \\
book/illustrations\_by              & {[}S{]} illustrated by {[}O{]}                   \\
religious\_text/religion       & {[}S{]} contains principles of {[}O{]} \\ 
spacecraft/manufacturer                & {[}S{]} was spacecraft developed by {[}O{]}     \\
\bottomrule
\end{tabular}
\caption{Table showing an example of textual contexts extracted for freebase predicates}
\label{table:example-dep}
\end{table}

\subsection{Generation of Special tokens}
To generate the special tokens for copy actions~(sec.~\ref{sec:copy-actions}) we run POS tagging on each of the input textual contexts\footnote{For the predicate textual contexts we run pos tagging on the original text not the lexicalized dependency path}. 
We replace every word in each textual context with a combination of its context id (e.g. \texttt{C1}) and its POS tag (e.g. \texttt{NOUN}). 
If the same POS tag appears multiple times in the textual context, it is given an additional id (e.g. \texttt{C1\_NOUN\_2}). 
If a word in the output question overlaps with a word in the input textual context, this word is replaced by its corresponding tag. \\
%
%
%
For sentence and word tokenization we use the Regex tokenizer from the NLTK toolkit~\cite{NLTK}, and for POS tagging and dependency parsing we use the Spacy\footnote{https://spacy.io/} implementation.
%
\section{Experiments}
%
\subsection{Zero-Shot Setups}~\label{sec:zeroshot-setup}
We develop three setups that follow the same procedure as~\cite{Levy17} for Zero-Shot relation extraction to evaluate how our model generalizes to: 1) unseen predicates, 2) unseen sub-types and 3) unseen obj-types. \\
For the unseen predicates setup we group all the samples in SimpleQuestions by the predicate of the input triple, and keep groups that contain at least 50 samples. 
Afterwards we randomly split those groups to 70\% train, 10\% valid and 20\% test mutual exclusive sets respectively.
This guarantees that if the predicate \texttt{fb:person/place\_of\_birth} for example shows during test time, the training and validation set will not contain any input triples having this predicate.
We repeat this process to create 10 cross validation folds, in our evaluation we report the mean and standard deviation results across those 10 folds. 
While doing this we make sure that the number of samples in each fold -- not only unique predicates -- follow the same 70\%, 30\%, 10\% distribution. 
We repeat the same process for the subject entity types and object entity types (answer types) individually.
Similarly, for example in the unseen object-type setup, the question \textit{"Which \textbf{artist} was born in Berlin?"} appearing in the test set means that, there is no question in the training set having an entity of type \textbf{\textit{artist}}. 
Table~\ref{tab:dataset-sizes} shows the mean number of samples, predicates, sub-types and obj-types across the 10 folds for each experiment setup. 
%

\setlength\extrarowheight{2pt} 

\begin{table}[t]
\footnotesize
\centering
\begin{tabular}{@{}l@{\hskip -0.08pt}llll@{}}
\toprule
                                                     &            & \textbf{Train} & \textbf{Valid} & \textbf{Test} \\ \midrule
\raisebox{-.95\normalbaselineskip}[0pt][0pt]{\rotatebox[origin=c]{90}{\textbf{\footnotesize{pred}}}}
& \# pred  & 169.4 & 24.2 & 48.4 \\
\multicolumn{1}{l}{}  & \# samples & 55566.7 & 7938.1 & 15876.2 \\
\multicolumn{1}{l}{}  & \% samples & 70.0 $\pm$ 2.77 & 10.0 $\pm$ 1.236 & 20.0 $\pm$ 2.12 \\
\midrule
\raisebox{-.95\normalbaselineskip}[0pt][0pt]{\rotatebox[origin=c]{90}{\textbf{\footnotesize{sub-types}}}}
& \# types  & 112.7 & 16.1 & 32.2 \\
\multicolumn{1}{l}{}  & \# samples & 60002.6 & 8571.8 & 17143.6 \\
\multicolumn{1}{l}{}  & \% samples & 70.0 $\pm$ 7.9 & 10.0 $\pm$ 3.6 & 20.0 $\pm$ 6.2 \\ \midrule

\raisebox{-.95\normalbaselineskip}[0pt][0pt]{\rotatebox[origin=c]{90}{\textbf{\footnotesize{obj-types}}}} 
& \# types  & 521.6 & 189.9 & 282.2 \\
\multicolumn{1}{l}{}  & \# samples & 57878.1 & 8268.3 & 16536.6 \\
\multicolumn{1}{l}{}  & \% samples & 70.0 $\pm$ 4.7 & 10.0 $\pm$ 2.5 & 20.0 $\pm$ 3.8 \\
\bottomrule
\end{tabular}
\caption{Dataset statistics across 10 folds for each experiment}
\label{tab:dataset-sizes}
\end{table}
\subsection{Baselines}~\label{sec:baselines}
\vspace{-15pt} 
%
\paragraph{\texttt{SELECT}}
is a baseline built from~\cite{Serban16} and adapted for the zero shot setup. 
During test time given a fact $F$, this baseline picks a fact $F_c$ from the training set and outputs the question that corresponds to it.
For evaluating unseen predicates, $F_c$ has the same answer type (obj-type) as $F$. 
And while evaluating unseen sub-types or obj-types, $F_c$ and $F$ have the same predicate.
%
\paragraph{\texttt{R-TRANSE}}
is an extension that we propose for \texttt{SELECT}. 
The input triple is encoded using the concatenation of the TransE embeddings of the subject, predicate and object. 
At test time, \texttt{R-TRANSE} picks a fact from the training set that is the closest to the input fact using cosine similarity and outputs the question that corresponds to it.
We provide two versions of this baseline:
\textbf{\texttt{R-TRANSE}} which indexes and retrieves raw questions with only a single placeholder for the subject label, such as in~\cite{Serban16}.
And \textbf{\texttt{R-TRANSE\textsubscript{copy}}} which indexes and retrieves questions using our copy actions mechanism (sec.~\ref{sec:copy-actions}). 
%
\paragraph{\texttt{IR}}
%
is an information retrieval baseline. Information retrieval has been used before as baseline for QG from text input~\cite{Rush15,Du17}. 
We rely on the textual context of each input triple as the search keyword for retrieval. 
First, the IR baseline encodes each question in the training set as a vector of TF-IDF weights~\cite{joachims_97} and then does dimensionality reduction through LSA~\cite{halko_svd_2011}.
At test time the textual context of the input triple is converted into a dense vector using the same process and then the question with the closest cosine distance to the input is retrieved. 
We provide two versions of this baseline: \textbf{\texttt{IR}} on raw text and \textbf{\texttt{IR\textsubscript{copy}}} on text with our placeholders for copy actions.

%
\paragraph{\texttt{Encoder-Decoder.}}
Finally, we compare our model to the Encoder-Decoder model with a single placeholder, the best performing model from~\cite{Serban16}.
We initialize the encoder with TransE embeddings and the decoder with GloVe word embeddings. 
Although this model was not originally built to generalize to unseen predicates and entity types, it has some generalization abilities represented in the encoded information in the pre-trained embeddings. 
Pretrained KB terms and word embeddings encode relations between entities or between words as translations in the vector space. 
Thus the model might be able to map new classes or predicates in the input fact to new words in the output question.
%
\begin{table*}[]
\footnotesize
\centering
\begin{tabular}{@{}ll|ccccccc@{}}
\toprule
& \textbf{Model} & \textbf{BLEU-1} & \textbf{BLEU-2} & \textbf{BLEU-3} & \textbf{BLEU-4} & \textbf{ROUGE\textsubscript{L}} & \textbf{METEOR} \\
\midrule[1.1pt]
\multirow{4}{*}{\raisebox{-3\normalbaselineskip}[0pt][0pt]{\rotatebox[origin=c]{90}{\textbf{\footnotesize{ Unseen Predicates}}}}}

&  SELECT & 46.81 $\pm$ 2.12 & 38.62 $\pm$ 1.78 & 31.26 $\pm$ 1.9 & 23.66 $\pm$ 2.22 & 52.04 $\pm$ 1.43 & 27.11 $\pm$ 0.74 \\
&  IR  & 48.43 $\pm$ 1.64 & 39.13 $\pm$ 1.34 & 31.4 $\pm$ 1.66 & 23.59 $\pm$ 2.36 & 52.88 $\pm$ 1.24 & 27.34 $\pm$ 0.55 \\
&  IR\textsubscript{COPY} & 48.22 $\pm$ 1.84 & 38.82 $\pm$ 1.5 & 31.01 $\pm$ 1.72 & 23.12 $\pm$ 2.24 & 52.72 $\pm$ 1.26 & 27.24 $\pm$ 0.57 \\
&  R-TRANSE & 49.09 $\pm$ 1.69 & 40.75 $\pm$ 1.42 & 33.4 $\pm$ 1.7 & 25.97 $\pm$ 2.22 & 54.07 $\pm$ 1.31 & 28.13 $\pm$ 0.54 \\
&  R-TRANSE\textsubscript{COPY}& 49.0 $\pm$ 1.76 & 40.63 $\pm$ 1.48 & 33.28 $\pm$ 1.74 & 25.87 $\pm$ 2.23 & 54.09 $\pm$ 1.35 & 28.12 $\pm$ 0.57 \\
&    Encoder-Decoder       & 58.92 $\pm$ 2.05 & 47.7 $\pm$ 1.62  & 38.18 $\pm$ 1.86 & 28.71 $\pm$ 2.35 & 59.12 $\pm$ 1.16 & 34.28 $\pm$ 0.54 \\
\cmidrule(l){2-8}
&  Our-Model      & 60.8  $\pm$ 1.52 & 49.8 $\pm$ 1.37  & 40.32 $\pm$ 1.92 & 30.76 $\pm$ 2.7  & 60.07 $\pm$ 0.9  & 35.34 $\pm$ 0.43 \\
& Our-Model\textsubscript{copy}    & \textbf{62.44} $\pm$ 1.85 & \textbf{50.62} $\pm$ 1.46 & \textbf{40.82} $\pm$ 1.77 & \textbf{31.1} $\pm$ 2.46  & \textbf{61.23} $\pm$ 1.2  & \textbf{36.24} $\pm$ 0.65 \\
\bottomrule
\end{tabular}
\caption{\small{Evaluation results of our model and all other baselines for the unseen predicate evaluation setup}}
\label{table:eval-pred}
\end{table*}
\begin{table}[t]
\footnotesize
\centering
\begin{tabular}{@{}ll|cc@{}}
\toprule
& \textbf{Model} & \textbf{BLEU-4} & \textbf{ROUGE\textsubscript{L}} \\
\midrule[1.1pt]
\multirow{4}{*}{\raisebox{-0.6\normalbaselineskip}[0pt][0pt]{\rotatebox[origin=c]{90}{\textbf{\footnotesize{Sub-Types}}}}}
&  R-TRANSE  & 32.41 $\pm$ 1.74 & 59.27 $\pm$ 0.92 \\
&  Encoder-Decoder         & 42.14 $\pm$ 2.05 & 68.95 $\pm$ 0.86 \\
\cmidrule(l){2-4}
& Our-Model                       & 42.13 $\pm$ 1.88 & 69.35 $\pm$ 0.9 \\
& Our-Model\textsubscript{copy}   & \textbf{42.2} $\pm$ 2.0 & \textbf{69.37} $\pm$ 1.0 \\
\midrule[1.2pt]
\multirow{4}{*}{\raisebox{-0.6\normalbaselineskip}[0pt][0pt]{\rotatebox[origin=c]{90}{\textbf{\footnotesize{Obj-Types}}}}}
&  R-TRANSE  & 30.59 $\pm$ 1.3 & 57.37 $\pm$ 1.17 \\
&  Encoder-Decoder          & 37.79 $\pm$ 2.65 & 65.69 $\pm$ 2.25 \\                      
\cmidrule(l){2-4}
& Our-Model        & 37.78 $\pm$ 2.02 & 65.51 $\pm$ 1.56 \\
& Our-Model\textsubscript{copy}            & \textbf{38.02} $\pm$ 1.9 & \textbf{66.24} $\pm$ 1.38 \\
\bottomrule
\end{tabular}
\caption{\small{Automatic evaluation of our model against selected baselines for unseen sub-types and obj-types}}
\label{table:eval-type}
\end{table}
\subsection{Training \& Implementation Details}
To train the neural network models we optimize the negative log-likelihood of the training data with respect to all the model parameters. For that we use the RMSProp optimization algorithm with a decreasing learning rate of~$0.001$, mini-batch size~$=200$, and clipping gradients with norms larger than $0.1$.
We use the same vocabulary for both the textual context encoders and the decoder outputs. We limit our vocabulary to the top $30,000$ words including the special tokens. 
For the word embeddings we chose GloVe~\cite{PenningtonSM14} pretrained embeddings of size $100$. 
We train TransE embeddings of size $H_k=200$, on the FB5M dataset~\cite{Bordes15} using the TransE model implementation from~\cite{Lin15}.
We set GRU hidden size of the decoder to $H_d=500$, and textual encoder to $H_c=200$.
The networks hyperparameters are set with respect to the final BLEU-4 score over the validation set.
%
All neural networks are implemented using Tensorflow~\cite{tensorflow}. %
All experiments and models source code are publicly available\footnote{\url{https://github.com/NAACL2018Anonymous/submission}} for the sake of reproducibility.
%
\subsection{Automatic Evaluation Metrics}
To evaluate the quality of the generated question, we compare the original labeled questions by human annotators to the ones generated by each variation of our model and the baselines.
We rely on a set of well established evaluation metrics for text generation: BLEU-1, BLEU-2, BLEU-3, BLEU-4~\cite{bleu}, METEOR~\cite{meteor} and ROUGE\textsubscript{L}~\cite{lin2004rouge}.
%
\\
\subsection{Human Evaluation}
Metrics for evaluating text generation such as BLEU and METEOR give an measure of how close the generated questions are to the target correct labels.
However, they might not be able to evaluate directly whether the predicate in the question was expressed in the text or not. 
%
%
%
Thus we run two further human evaluations to directly measure the following. \\
\textbf{\textit{Predicate identification}}: annotators were asked to indicate whether the generated question contains the given predicate in the fact or not, either directly or implicitly. \\
\textbf{\textit{Naturalness}}: following~\cite{Ngomo13}, we measure  the comprehensibility and readability of the generated questions.
Each annotator was asked to rate each generated question using a scale from $1$ to $5$, where: 
(5) perfectly clear and natural, 
(3) artificial but understandable, and (1) completely not understandable.
We run our studies on $100$ randomly sampled input facts alongside with their corresponding generated questions by each of the systems using the help of $4$ annotators.
%
%
\section{Results \& Discussion}
\paragraph{Automatic Evaluation} Table~\ref{table:eval-pred} shows results of our model compared to all other baselines across all evaluation metrics. 
%
Our that encodes the KB fact and textual contexts achieves a significant enhancement over all the baselines in all evaluation metrics, with $\mathbf{+2.04}$ BLEU-4 score than the Encoder-Decoder baseline.
Incorporating the part-of-speech copy actions further improves this enhancement to reach \textbf{$\mathbf{+2.39}$} BLEU-4 points. \\
%
Among all baselines, the Encoder-Decoder baseline and the R-TRANSE baseline performed the best. 
This shows that TransE embeddings encode  intra-predicates information and intra-class-types information to a great extent, and can generalize to some extent to unseen predicates and class types.
Similar patterns can be seen in the evaluation on unseen sub-types and obj-types (Table~\ref{table:eval-type}). Our model with copy actions was able to outperform all the other systems.
Majority of systems have reported a significantly higher BLEU-4 scores in these two tasks than when generalizing to unseen predicates ($+12$ and $+8$ BLEU-4 points respectively). 
This indicates that these tasks are relatively easier and hence our models achieve relatively smaller enhancements over the baselines. 
%
\begin{table}[t]
\small
\centering
\begin{tabular}{@{}l@{\hskip -3pt}cc@{}}
\toprule
\textbf{Model} & \textbf{$\%$ Pred. Identified}    & \textbf{Natural.} \\ \midrule
Encoder-Decoder                 & $6$                &   $3.14$          \\
Our-Model (No Copy)         & $6$                &   $2.72$          \\
Our-Model\textsubscript{copy} (Types context) & $\mathbf{37}$      &    $\mathbf{3.21}$         \\
Our-Model\textsubscript{copy} (All contexts)       & $\mathbf{46}$ & $2.61$           \\ \midrule
\end{tabular}
\caption{results of Human evaluation on \% of predicates identified and naturalness 0-5}
\label{table:human-eval}
\end{table}

\begin{table}[t]
\centering
\footnotesize
\begin{tabularx}{1.05\linewidth}{l|lX} 
\toprule
1&\textbf{Reference} & \textbf{what language is spoken in the tv show three sheets?}                                      \\
&\textbf{Enc-Dec.}   & in what \textbf{language} is three sheets in?                                    \\
&\textbf{Our-Model}        & what the the player is the three sheets?                           \\
&\textbf{Our-Model\textsubscript{Copy}}      & what is the \textbf{language} of three sheets? \\\midrule
2&\textbf{Reference} & \textbf{how is roosevelt in Africa classified?                        } \\
&\textbf{Enc-Dec.}   & what is the name of a roosevelt in Africa?                    \\
&\textbf{Our-Model}        & what is the name of the movie roosevelt in Africa?               \\
&\textbf{Our-Model\textsubscript{Copy}}     & what is a \textbf{genre} of roosevelt in Africa?                          \\ \midrule
3&\textbf{Reference} & \textbf{where can 5260 philvéron be found}?                              \\
&\textbf{Enc-Dec.}   & what is a release some that 5260 philvéron wrote?                \\
&\textbf{Our-Model}        & what is the name of an artist 5260 philvéron?                    \\
&\textbf{Our-Model\textsubscript{Copy}}     & which \textbf{star system} contains the star system body 5260 philvéron?  \\ \midrule
4&\textbf{Reference} & \textbf{which university did ezra cornell create}?             \\
&\textbf{Enc-Dec.}   & which films are part of ezra cornell?             \\
&\textbf{Our-Model}        & what is a position of ezra cornell?                              \\
&\textbf{Our-Model\textsubscript{Copy}}     & what \textit{founded} the name of a university that ezra cornell \textbf{founded}? \\ 

\bottomrule
\end{tabularx}
\caption{Examples of generated questions from different systems in comparison}
\label{table:results-example}
\end{table}
\paragraph{Human Evaluation}
Table~\ref{table:human-eval} shows how different variations of our system can express the unseen predicate in the target question with comparison to the Encoder-Decoder baseline. \\
Our proposed copy actions have scored a significant enhancement in the identification of unseen predicates with up to $\mathbf{+40\%}$ more than best performing baseline and our model version without the copy actions. \\
%
By examining some of the generated questions~(Table~\ref{table:results-example}) we see that models without copy actions can generalize to unseen predicates that only have a very similar freebase predicate in the training set. For example \texttt{fb:tv\_program/language} and \texttt{fb:film/language}, if one of those predicates exists in the training set the model can use the same questions for the other during test time. \\ 
Copy actions from the sub-type and the obj-type textual contexts can generalize to a great extent to unseen predicates because of the overlap between the predicate and the object type in many questions (Example 2 Table~\ref{table:results-example}). %
Adding the predicate context to our model has enhanced model performance for expressing unseen predicates by $+9\%$ (Table~\ref{table:human-eval}).
However we can see that it has affected the naturalness of the question. 
The post processing step does not take into consideration that some verbs and prepositions do not fit in the sentence structure, or that some words are already existing in the question words (Example 4 Table~\ref{table:results-example}).
This does not happen as much when having copy actions from the sub-type and the obj-type contexts because they are mainly formed of nouns which are more interchangeable than verbs or prepositions.
A post-processing step to reform the question instead of direct copying from the input source is considered in our future work.
%
\section{Conclusion}
%
In this paper we presented a new neural model for question generation from knowledge bases, with a main focus on predicates, subject types or object types that were not seen at the training phase (Zero-Shot Question Generation). %
Our model is based on an encoder-decoder architecture that leverages textual contexts of triples, two attention layers for triples and textual contexts and finally a part-of-speech copy action mechanism. %
Our method exhibits significantly better results for Zero-Shot QG than a set of strong baselines including the state-of-the-art question generation from KB. %
Additionally, a complimentary human evaluation, helps in showing that the improvement brought by our part-of-speech copy action mechanism is even more significant than what the automatic evaluation suggests.
The source code and the collected textual contexts are provided for the community: \url{https://github.com/NAACL2018Anonymous/submission}
\bibliography{naaclhlt2018}

\begin{thebibliography}{}
\expandafter\ifx\csname natexlab\endcsname\relax\def\natexlab#1{#1}\fi

\bibitem[{Abadi et~al.(2015)Abadi, Agarwal, Barham, Brevdo, Chen, Citro,
  Corrado, Davis, Dean, Devin, Ghemawat, Goodfellow, Harp, Irving, Isard, Jia,
  Jozefowicz, Kaiser, Kudlur, Levenberg, Man\'{e}, Monga, Moore, Murray, Olah,
  Schuster, Shlens, Steiner, Sutskever, Talwar, Tucker, Vanhoucke, Vasudevan,
  Vi\'{e}gas, Vinyals, Warden, Wattenberg, Wicke, Yu, and Zheng}]{tensorflow}
Mart\'{\i}n Abadi, Ashish Agarwal, Paul Barham, Eugene Brevdo, Zhifeng Chen,
  Craig Citro, Greg~S. Corrado, Andy Davis, Jeffrey Dean, Matthieu Devin,
  Sanjay Ghemawat, Ian Goodfellow, Andrew Harp, Geoffrey Irving, Michael Isard,
  Yangqing Jia, Rafal Jozefowicz, Lukasz Kaiser, Manjunath Kudlur, Josh
  Levenberg, Dan Man\'{e}, Rajat Monga, Sherry Moore, Derek Murray, Chris Olah,
  Mike Schuster, Jonathon Shlens, Benoit Steiner, Ilya Sutskever, Kunal Talwar,
  Paul Tucker, Vincent Vanhoucke, Vijay Vasudevan, Fernanda Vi\'{e}gas, Oriol
  Vinyals, Pete Warden, Martin Wattenberg, Martin Wicke, Yuan Yu, and Xiaoqiang
  Zheng. 2015.
\newblock \href{https://www.tensorflow.org/}{{TensorFlow}: Large-scale machine
  learning on heterogeneous systems}.
\newblock Software available from tensorflow.org.
\newblock \url{https://www.tensorflow.org/}.

\bibitem[{Bahdanau et~al.(2014)Bahdanau, Cho, and Bengio}]{Bahdanau14}
Dzmitry Bahdanau, Kyunghyun Cho, and Yoshua Bengio. 2014.
\newblock \href{http://arxiv.org/abs/1409.0473}{Neural machine translation by
  jointly learning to align and translate}.
\newblock {\em CoRR\/} abs/1409.0473.
\newblock \url{http://arxiv.org/abs/1409.0473}.

\bibitem[{Bird(2006)}]{NLTK}
Steven Bird. 2006.
\newblock \href{http://aclweb.org/anthology/P06-4018}{{NLTK:} the natural
  language toolkit}.
\newblock In {\em {ACL} 2006, 21st International Conference on Computational
  Linguistics and 44th Annual Meeting of the Association for Computational
  Linguistics, Proceedings of the Conference, Sydney, Australia, 17-21 July
  2006\/}.
\newblock \url{http://aclweb.org/anthology/P06-4018}.

\bibitem[{Bordes et~al.(2015)Bordes, Usunier, Chopra, and Weston}]{Bordes15}
Antoine Bordes, Nicolas Usunier, Sumit Chopra, and Jason Weston. 2015.
\newblock \href{http://arxiv.org/abs/1506.02075}{Large-scale simple question
  answering with memory networks}.
\newblock {\em CoRR\/} abs/1506.02075.
\newblock \url{http://arxiv.org/abs/1506.02075}.

\bibitem[{Cho et~al.(2014)Cho, van Merrienboer, G{\"{u}}l{\c{c}}ehre, Bougares,
  Schwenk, and Bengio}]{Cho2014}
Kyunghyun Cho, Bart van Merrienboer, {\c{C}}aglar G{\"{u}}l{\c{c}}ehre, Fethi
  Bougares, Holger Schwenk, and Yoshua Bengio. 2014.
\newblock Learning phrase representations using {RNN} encoder-decoder for
  statistical machine translation.
\newblock {\em CoRR\/} abs/1406.1078.

\bibitem[{Denkowski and Lavie(2014)}]{meteor}
Michael~J. Denkowski and Alon Lavie. 2014.
\newblock \href{http://aclweb.org/anthology/W/W14/W14-3348.pdf}{Meteor
  universal: Language specific translation evaluation for any target language}.
\newblock In {\em Proceedings of the Ninth Workshop on Statistical Machine
  Translation, WMT@ACL 2014, June 26-27, 2014, Baltimore, Maryland, {USA}\/}.
  pages 376--380.
\newblock \url{http://aclweb.org/anthology/W/W14/W14-3348.pdf}.

\bibitem[{Dong et~al.(2017)Dong, Mallinson, Reddy, and Lapata}]{Dong17}
Li~Dong, Jonathan Mallinson, Siva Reddy, and Mirella Lapata. 2017.
\newblock \href{https://aclanthology.info/papers/D17-1091/d17-1091}{Learning to
  paraphrase for question answering}.
\newblock In {\em Proceedings of the 2017 Conference on Empirical Methods in
  Natural Language Processing, {EMNLP} 2017, Copenhagen, Denmark, September
  9-11, 2017\/}. pages 875--886.
\newblock \url{https://aclanthology.info/papers/D17-1091/d17-1091}.

\bibitem[{Du et~al.(2017)Du, Shao, and Cardie}]{Du17}
Xinya Du, Junru Shao, and Claire Cardie. 2017.
\newblock \href{https://doi.org/10.18653/v1/P17-1123}{Learning to ask: Neural
  question generation for reading comprehension}.
\newblock In {\em Proceedings of the 55th Annual Meeting of the Association for
  Computational Linguistics, {ACL} 2017, Vancouver, Canada, July 30 - August 4,
  Volume 1: Long Papers\/}. pages 1342--1352.
\newblock \url{https://doi.org/10.18653/v1/P17-1123}.

\bibitem[{Evens and Michael(2006)}]{evens2006}
Martha Evens and Joel Michael. 2006.
\newblock One-on-one tutoring by humans and machines.
\newblock {\em Computer Science Department, Illinois Institute of Technology\/}
  .

\bibitem[{Fader et~al.(2011)Fader, Soderland, and Etzioni}]{Fader11}
Anthony Fader, Stephen Soderland, and Oren Etzioni. 2011.
\newblock \href{http://www.aclweb.org/anthology/D11-1142}{Identifying relations
  for open information extraction}.
\newblock In {\em Proceedings of the 2011 Conference on Empirical Methods in
  Natural Language Processing, {EMNLP} 2011, 27-31 July 2011, John McIntyre
  Conference Centre, Edinburgh, UK, {A} meeting of SIGDAT, a Special Interest
  Group of the {ACL}\/}. pages 1535--1545.
\newblock \url{http://www.aclweb.org/anthology/D11-1142}.

\bibitem[{Graesser et~al.(2004)Graesser, Lu, Jackson, Mitchell, Ventura, Olney,
  and Louwerse}]{graesser2004}
Arthur~C Graesser, Shulan Lu, George~Tanner Jackson, Heather~Hite Mitchell,
  Mathew Ventura, Andrew Olney, and Max~M Louwerse. 2004.
\newblock Autotutor: A tutor with dialogue in natural language.
\newblock {\em Behavior Research Methods\/} 36(2):180--192.

\bibitem[{Gu et~al.(2016)Gu, Lu, Li, and Li}]{Gu16}
Jiatao Gu, Zhengdong Lu, Hang Li, and Victor O.~K. Li. 2016.
\newblock \href{http://aclweb.org/anthology/P/P16/P16-1154.pdf}{Incorporating
  copying mechanism in sequence-to-sequence learning}.
\newblock In {\em Proceedings of the 54th Annual Meeting of the Association for
  Computational Linguistics, {ACL} 2016, August 7-12, 2016, Berlin, Germany,
  Volume 1: Long Papers\/}.
\newblock \url{http://aclweb.org/anthology/P/P16/P16-1154.pdf}.

\bibitem[{G{\"{u}}l{\c{c}}ehre et~al.(2016)G{\"{u}}l{\c{c}}ehre, Ahn,
  Nallapati, Zhou, and Bengio}]{Gulcehre16}
{\c{C}}aglar G{\"{u}}l{\c{c}}ehre, Sungjin Ahn, Ramesh Nallapati, Bowen Zhou,
  and Yoshua Bengio. 2016.
\newblock \href{http://aclweb.org/anthology/P/P16/P16-1014.pdf}{Pointing the
  unknown words}.
\newblock In {\em Proceedings of the 54th Annual Meeting of the Association for
  Computational Linguistics, {ACL} 2016, August 7-12, 2016, Berlin, Germany,
  Volume 1: Long Papers\/}.
\newblock \url{http://aclweb.org/anthology/P/P16/P16-1014.pdf}.

\bibitem[{Halko et~al.(2011)Halko, Martinsson, and Tropp}]{halko_svd_2011}
Nathan Halko, Per{-}Gunnar Martinsson, and Joel~A. Tropp. 2011.
\newblock \href{https://doi.org/10.1137/090771806}{Finding structure with
  randomness: Probabilistic algorithms for constructing approximate matrix
  decompositions}.
\newblock {\em {SIAM} Review\/} 53(2):217--288.
\newblock \url{https://doi.org/10.1137/090771806}.

\bibitem[{Heilman and Smith(2010)}]{HeilmanS10}
Michael Heilman and Noah~A. Smith. 2010.
\newblock \href{http://www.aclweb.org/anthology/N10-1086}{Good question!
  statistical ranking for question generation}.
\newblock In {\em Human Language Technologies: Conference of the North American
  Chapter of the Association of Computational Linguistics, Proceedings, June
  2-4, 2010, Los Angeles, California, {USA}\/}. pages 609--617.
\newblock \url{http://www.aclweb.org/anthology/N10-1086}.

\bibitem[{Joachims(1997)}]{joachims_97}
Thorsten Joachims. 1997.
\newblock A probabilistic analysis of the rocchio algorithm with {TFIDF} for
  text categorization.
\newblock In {\em Proceedings of the Fourteenth International Conference on
  Machine Learning {(ICML} 1997), Nashville, Tennessee, USA, July 8-12,
  1997\/}. pages 143--151.

\bibitem[{Khapra et~al.(2017)Khapra, Raghu, Joshi, and Reddy}]{Khapra17}
Mitesh~M. Khapra, Dinesh Raghu, Sachindra Joshi, and Sathish Reddy. 2017.
\newblock \href{https://aclanthology.info/pdf/E/E17/E17-1036.pdf}{Generating
  natural language question-answer pairs from a knowledge graph using a {RNN}
  based question generation model}.
\newblock In {\em Proceedings of the 15th Conference of the European Chapter of
  the Association for Computational Linguistics, {EACL} 2017, Valencia, Spain,
  April 3-7, 2017, Volume 1: Long Papers\/}. pages 376--385.
\newblock \url{https://aclanthology.info/pdf/E/E17/E17-1036.pdf}.

\bibitem[{Kiros et~al.(2014)Kiros, Salakhutdinov, and Zemel}]{Kiros14}
Ryan Kiros, Ruslan Salakhutdinov, and Richard~S. Zemel. 2014.
\newblock \href{http://arxiv.org/abs/1411.2539}{Unifying visual-semantic
  embeddings with multimodal neural language models}.
\newblock {\em CoRR\/} abs/1411.2539.
\newblock \url{http://arxiv.org/abs/1411.2539}.

\bibitem[{Lebret et~al.(2016)Lebret, Grangier, and Auli}]{lebret2016}
R{\'{e}}mi Lebret, David Grangier, and Michael Auli. 2016.
\newblock \href{http://aclweb.org/anthology/D/D16/D16-1128.pdf}{Neural text
  generation from structured data with application to the biography domain}.
\newblock In {\em Proceedings of the 2016 Conference on Empirical Methods in
  Natural Language Processing, {EMNLP} 2016, Austin, Texas, USA, November 1-4,
  2016\/}. pages 1203--1213.
\newblock \url{http://aclweb.org/anthology/D/D16/D16-1128.pdf}.

\bibitem[{Levy et~al.(2017)Levy, Seo, Choi, and Zettlemoyer}]{Levy17}
Omer Levy, Minjoon Seo, Eunsol Choi, and Luke Zettlemoyer. 2017.
\newblock \href{https://doi.org/10.18653/v1/K17-1034}{Zero-shot relation
  extraction via reading comprehension}.
\newblock In {\em Proceedings of the 21st Conference on Computational Natural
  Language Learning (CoNLL 2017), Vancouver, Canada, August 3-4, 2017\/}. pages
  333--342.
\newblock \url{https://doi.org/10.18653/v1/K17-1034}.

\bibitem[{Lin(2004)}]{lin2004rouge}
Chin-Yew Lin. 2004.
\newblock Rouge: A package for automatic evaluation of summaries.
\newblock In {\em Text summarization branches out: Proceedings of the ACL-04
  workshop\/}. Barcelona, Spain, volume~8.

\bibitem[{Lin et~al.(2015)Lin, Liu, Sun, Liu, and Zhu}]{Lin15}
Yankai Lin, Zhiyuan Liu, Maosong Sun, Yang Liu, and Xuan Zhu. 2015.
\newblock
  \href{http://www.aaai.org/ocs/index.php/AAAI/AAAI15/paper/view/9571}{Learning
  entity and relation embeddings for knowledge graph completion}.
\newblock In {\em Proceedings of the Twenty-Ninth {AAAI} Conference on
  Artificial Intelligence, January 25-30, 2015, Austin, Texas, {USA.}\/}. pages
  2181--2187.
\newblock \url{http://www.aaai.org/ocs/index.php/AAAI/AAAI15/paper/view/9571}.

\bibitem[{Luong et~al.(2015)Luong, Sutskever, Le, Vinyals, and
  Zaremba}]{Luong15}
Thang Luong, Ilya Sutskever, Quoc~V. Le, Oriol Vinyals, and Wojciech Zaremba.
  2015.
\newblock \href{http://aclweb.org/anthology/P/P15/P15-1002.pdf}{Addressing the
  rare word problem in neural machine translation}.
\newblock In {\em Proceedings of the 53rd Annual Meeting of the Association for
  Computational Linguistics and the 7th International Joint Conference on
  Natural Language Processing of the Asian Federation of Natural Language
  Processing, {ACL} 2015, July 26-31, 2015, Beijing, China, Volume 1: Long
  Papers\/}. pages 11--19.
\newblock \url{http://aclweb.org/anthology/P/P15/P15-1002.pdf}.

\bibitem[{Mintz et~al.(2009)Mintz, Bills, Snow, and Jurafsky}]{mintz_2009}
Mike Mintz, Steven Bills, Rion Snow, and Daniel Jurafsky. 2009.
\newblock \href{http://www.aclweb.org/anthology/P09-1113}{Distant supervision
  for relation extraction without labeled data}.
\newblock In {\em {ACL} 2009, Proceedings of the 47th Annual Meeting of the
  Association for Computational Linguistics and the 4th International Joint
  Conference on Natural Language Processing of the AFNLP, 2-7 August 2009,
  Singapore\/}. pages 1003--1011.
\newblock \url{http://www.aclweb.org/anthology/P09-1113}.

\bibitem[{Ngomo et~al.(2013)Ngomo, B{\"{u}}hmann, Unger, Lehmann, and
  Gerber}]{Ngomo13}
Axel{-}Cyrille~Ngonga Ngomo, Lorenz B{\"{u}}hmann, Christina Unger, Jens
  Lehmann, and Daniel Gerber. 2013.
\newblock \href{http://dl.acm.org/citation.cfm?id=2488473}{Sorry, i don't speak
  {SPARQL:} translating {SPARQL} queries into natural language}.
\newblock In {\em 22nd International World Wide Web Conference, {WWW} '13, Rio
  de Janeiro, Brazil, May 13-17, 2013\/}. pages 977--988.
\newblock \url{http://dl.acm.org/citation.cfm?id=2488473}.

\bibitem[{Papineni et~al.(2002)Papineni, Roukos, Ward, and Zhu}]{bleu}
Kishore Papineni, Salim Roukos, Todd Ward, and Wei{-}Jing Zhu. 2002.
\newblock \href{http://www.aclweb.org/anthology/P02-1040.pdf}{Bleu: a method
  for automatic evaluation of machine translation}.
\newblock In {\em Proceedings of the 40th Annual Meeting of the Association for
  Computational Linguistics, July 6-12, 2002, Philadelphia, PA, {USA.}\/}.
  pages 311--318.
\newblock \url{http://www.aclweb.org/anthology/P02-1040.pdf}.

\bibitem[{Pennington et~al.(2014)Pennington, Socher, and
  Manning}]{PenningtonSM14}
Jeffrey Pennington, Richard Socher, and Christopher~D. Manning. 2014.
\newblock \href{http://aclweb.org/anthology/D/D14/D14-1162.pdf}{Glove: Global
  vectors for word representation}.
\newblock In {\em Proceedings of the 2014 Conference on Empirical Methods in
  Natural Language Processing, {EMNLP} 2014, October 25-29, 2014, Doha, Qatar,
  {A} meeting of SIGDAT, a Special Interest Group of the {ACL}\/}. pages
  1532--1543.
\newblock \url{http://aclweb.org/anthology/D/D14/D14-1162.pdf}.

\bibitem[{Riedel et~al.(2010)Riedel, Yao, and McCallum}]{riedel_2010}
Sebastian Riedel, Limin Yao, and Andrew McCallum. 2010.
\newblock \href{https://doi.org/10.1007/978-3-642-15939-8_10}{Modeling
  relations and their mentions without labeled text}.
\newblock In {\em Machine Learning and Knowledge Discovery in Databases,
  European Conference, {ECML} {PKDD} 2010, Barcelona, Spain, September 20-24,
  2010, Proceedings, Part {III}\/}. pages 148--163.
\newblock \url{https://doi.org/10.1007/978-3-642-15939-8_10}.

\bibitem[{Rush et~al.(2015)Rush, Chopra, and Weston}]{Rush15}
Alexander~M. Rush, Sumit Chopra, and Jason Weston. 2015.
\newblock \href{http://aclweb.org/anthology/D/D15/D15-1044.pdf}{A neural
  attention model for abstractive sentence summarization}.
\newblock In {\em Proceedings of the 2015 Conference on Empirical Methods in
  Natural Language Processing, {EMNLP} 2015, Lisbon, Portugal, September 17-21,
  2015\/}. pages 379--389.
\newblock \url{http://aclweb.org/anthology/D/D15/D15-1044.pdf}.

\bibitem[{Serban et~al.(2016)Serban, Garc{\'{\i}}a{-}Dur{\'{a}}n,
  G{\"{u}}l{\c{c}}ehre, Ahn, Chandar, Courville, and Bengio}]{Serban16}
Iulian~Vlad Serban, Alberto Garc{\'{\i}}a{-}Dur{\'{a}}n, {\c{C}}aglar
  G{\"{u}}l{\c{c}}ehre, Sungjin Ahn, Sarath Chandar, Aaron~C. Courville, and
  Yoshua Bengio. 2016.
\newblock \href{http://aclweb.org/anthology/P/P16/P16-1056.pdf}{Generating
  factoid questions with recurrent neural networks: The 30m factoid
  question-answer corpus}.
\newblock In {\em Proceedings of the 54th Annual Meeting of the Association for
  Computational Linguistics, {ACL} 2016, August 7-12, 2016, Berlin, Germany,
  Volume 1: Long Papers\/}.
\newblock \url{http://aclweb.org/anthology/P/P16/P16-1056.pdf}.

\bibitem[{Seyler et~al.(2015)Seyler, Yahya, and Berberich}]{SeylerYB15}
Dominic Seyler, Mohamed Yahya, and Klaus Berberich. 2015.
\newblock \href{https://doi.org/10.1145/2740908.2742722}{Generating quiz
  questions from knowledge graphs}.
\newblock In {\em Proceedings of the 24th International Conference on World
  Wide Web Companion, {WWW} 2015, Florence, Italy, May 18-22, 2015 - Companion
  Volume\/}. pages 113--114.
\newblock \url{https://doi.org/10.1145/2740908.2742722}.

\bibitem[{Shang et~al.(2015)Shang, Lu, and Li}]{Shang15}
Lifeng Shang, Zhengdong Lu, and Hang Li. 2015.
\newblock \href{http://aclweb.org/anthology/P/P15/P15-1152.pdf}{Neural
  responding machine for short-text conversation}.
\newblock In {\em Proceedings of the 53rd Annual Meeting of the Association for
  Computational Linguistics and the 7th International Joint Conference on
  Natural Language Processing of the Asian Federation of Natural Language
  Processing, {ACL} 2015, July 26-31, 2015, Beijing, China, Volume 1: Long
  Papers\/}. pages 1577--1586.
\newblock \url{http://aclweb.org/anthology/P/P15/P15-1152.pdf}.

\bibitem[{Socher et~al.(2013)Socher, Ganjoo, Manning, and Ng}]{Socher13}
Richard Socher, Milind Ganjoo, Christopher~D. Manning, and Andrew~Y. Ng. 2013.
\newblock Zero-shot learning through cross-modal transfer.
\newblock In {\em Advances in Neural Information Processing Systems 26: 27th
  Annual Conference on Neural Information Processing Systems 2013. Proceedings
  of a meeting held December 5-8, 2013, Lake Tahoe, Nevada, United States.\/}.
  pages 935--943.

\bibitem[{Sutskever et~al.(2014)Sutskever, Vinyals, and Le}]{Sutskever14}
Ilya Sutskever, Oriol Vinyals, and Quoc~V. Le. 2014.
\newblock Sequence to sequence learning with neural networks.
\newblock In {\em Advances in Neural Information Processing Systems 27: Annual
  Conference on Neural Information Processing Systems 2014, December 8-13 2014,
  Montreal, Quebec, Canada\/}. pages 3104--3112.

\end{thebibliography}


\begin{thebibliography}{}
\expandafter\ifx\csname natexlab\endcsname\relax\def\natexlab#1{#1}\fi

\end{thebibliography}
\bibliographystyle{acl_natbib}
\end{document}